\documentclass[sigconf]{acmart}

\AtBeginDocument{%
  }

\acmConference[Conference]{Make sure to enter the correct
  conference title from your rights confirmation email}

\usepackage{multirow, multicol,adjustbox}
\usepackage{comment}
\usepackage{algorithm}
\usepackage{algpseudocode}

\begin{document}
\title{Advancing climate model interpretability: Feature attribution for Arctic melt anomalies}

\author{Tolulope Ale}
\affiliation{
    \institution{Institute for Harnessing Data and} 
\text{Model Revolution in the Polar Regions } 
\text{University of Maryland, Baltimore County}
\city{Baltimore}
\state{MD}
\country{USA}}
\email{tale2@umbc.edu}

\author{Nicole-Jeanne Schlegel}
\affiliation{
\institution{National Oceanic and Atmospheric}
\text{Administration/OAR, Geophysical} 
    \text{Fluid Dynamic Laboratory}\\
\city{Princeton}
\state{NJ}
\country{USA}} 
\email{nicole.schlegel@noaa.gov}

\author{Vandana P. Janeja}
\affiliation{
\institution{Institute for Harnessing Data and} 
\text{Model Revolution in the Polar Regions } 
\text{University of Maryland, Baltimore County}
\city{Baltimore}
\state{MD}
\country{USA}} 
\email{vjaneja@umbc.edu}

\renewcommand{\shortauthors}{Ale et al.}

\begin{abstract}  
The focus of our work is improving the interpretability of anomalies in climate models and advancing our understanding of Arctic melt dynamics.

The Arctic and Antarctic ice sheets are experiencing rapid surface melting and increased freshwater runoff, contributing significantly to global sea level rise. Understanding the mechanisms driving snowmelt in these regions is crucial. ERA5, a widely used reanalysis dataset in polar climate studies, offers extensive climate variables and global data assimilation. However, its snowmelt model employs an energy imbalance approach that may oversimplify the complexity of surface melt. In contrast, the Glacier Energy and Mass Balance (GEMB) model incorporates additional physical processes, such as snow accumulation, firn densification, and meltwater percolation/refreezing, providing a more detailed representation of surface melt dynamics.  

In this research, we focus on analyzing surface snowmelt dynamics of the Greenland Ice Sheet using feature attribution for anomalous melt events in ERA5 and GEMB models. We present a novel unsupervised attribution method leveraging counterfactual explanation method to analyze detected anomalies in ERA5 and GEMB. Our anomaly detection results are validated using MEaSUREs ground-truth data, and the attributions are evaluated against established feature ranking methods, including XGBoost, Shapley values, and Random Forest.  

Our attribution framework identifies the physics behind each model and the climate features driving melt anomalies. 
These findings demonstrate the utility of our attribution method in enhancing the interpretability of anomalies in climate models and advancing our understanding of Arctic melt dynamics.  
\end{abstract}  

\maketitle 
\section{Introduction}
The Arctic and Antarctic ice sheets are undergoing accelerated surface melting, contributing significantly to global sea level rise and threatening coastal communities worldwide. These regions are particularly sensitive to atmospheric warming, with the Arctic experiencing warming approximately 3.8 times faster than the global average since 1979 \cite{rantanen2022arctic}. Surface melt, driven by complex interactions between climate variables, has emerged as a dominant factor in ice mass loss, underscoring the need for accurate modeling to project future climate impacts \cite{IPCC_2023, NOAA}.

The Greenland Ice Sheet, as a key contributor to sea level rise, presents a unique opportunity to investigate snowmelt processes under changing climate conditions. However, modeling snowmelt in this region is highly challenging due to the complexity of cryosph-
eric processes and uncertainties in existing models \cite{bamber2022ice, holube2022sources}. Identifying and attributing the primary drivers of anomalous melt events is critical to understanding these dynamics and improving the predictive capabilities of snowmelt models.

The ERA5 reanalysis dataset \cite{hersbach2020era5} and the Glacier Energy and Mass Balance (GEMB) \cite{gardner2023glacier} model, represent distinct approaches to modeling snowmelt. ERA5, widely used in climate studies, offers comprehensive atmospheric data but its snowmelt model relies on an energy imbalance framework, which simplifies cryospheric processes. It has limitations, such as overestimations of snow water equivalent \cite{kouki2023evaluation} and the assignment of a constant snow mass of 10 m to glacier grid points \cite{munoz2021era5}. In contrast, GEMB melt model integrates additional physical processes, including snow accumulation, firn densification, and meltwater refreezing, providing a more nuanced understanding of melt dynamics \cite{gardner2023glacier}. However, GEMB does not account for atmosphere-glacier feedback mechanisms, which can limit its applicability in certain contexts.

Given that the GEMB melt model is forced by ERA5 reanalysis data, this research aims to identify and analyze anomalous patterns that emerge when substituting the ERA5 snowmelt model with the GEMB melt model in a multivariate analysis. Using the Cluster-LSTM-VAE framework for anomaly detection \cite{ale2024harnessing}, we identify anomalous patterns. A counterfactual explanation method is employed to provide instance-level feature attribution of anomalies to a specific climate variable. This approach offers insights into the physics of surface melt in each model. We evaluated the attribution method against established feature importance techniques, such as Random Forest, XGBoost, and SHAP values, to ensure robustness and reliability, and we also evaluated the pattern of anomalies detected against observed ground truth.

The key contributions of this research are as follows: 
\begin{itemize} 
\item Developing an instance-level attribution mechanism for anomalous melt in unsupervised stochastic learning models. 
\item Analyzing and comparing anomalous patterns identified within the ERA5 and GEMB melt models over the Greenland Ice Sheet. 
\item Providing insights into the distinct representations of cryos-\\
pheric melt dynamics by these models.
\end{itemize}

This research contributes to improved understanding and modeling of Arctic melt processes.

The following sections are structured as follows: Section 2 reviews related work on anomaly detection and attribution in climate modeling. Section 3 discusses the Cluster-LSTM-VAE algorithm and our methodology for attribution, while Section 4 presents the experimental setup and results. Finally, Section 5 concludes with a summary and future research directions.


\section{Related Work}
Accurate modeling of the Greenland Ice Sheet requires understanding the interplay of highly interconnected climate variables. Traditional univariate anomaly detection methods, while effective for specific analysis, fail to capture critical insights arising from interactions between variables \cite{flach2017multivariate}. Multivariate analysis addresses this limitation by identifying anomalies resulting from the interaction of multiple variables.

Since no real-world scenario happens in isolation, coupled with the recent advancements in computational techniques, the multivariate analysis of extreme climate events has become significant to increase our understanding. The Greenland Ice Sheet exhibits strong interaction among multiple variables. For instance, the significant mass loss and melting in 2019 were partially due to low snowfall during the preceding winter \cite{nsidc2019} and an unprecedented rainfall at the summit of Greenland ice sheet in 2021 triggered a three-day melt episode \cite{overland2022arctic}, demonstrating how interconnected variables drive surface snow melt. Multivariate analysis offers further insights into these dynamics. For example, Kao et al. (2009) \cite{kao2009motivating} showcased the utility of copula-based methods in modeling complex dependencies between variables in hydrometeorological data, enabling the detection of compound anomalies often missed by traditional univariate methods. Additionally, Flach et al. (2017) \cite{flach2017multivariate} identifies workflow for detecting anomalous patterns in multivariate Earth observation by highlighting the importance of feature extraction steps.  

G Pang et al. (2021) \cite{pang2021deep} classify anomaly detection using deep learning into feature extraction, representation learning, and end-to-end detection method. Representation learning techniques, such as Variational Autoencoders (VAEs), model complex data distributions in an unsupervised approach, making it suitable for anomaly detection in multivariate climate datasets to capture intricate relationships between variables.
Deep learning methods have significantly advanced multivariate anomaly detection. For example, variational autoencoders (VAEs) \cite{an2015variational, sinha2020variational}, transformers \cite{tuli2022tranad, wang2022variational}, long short-term memory (LSTM) networks \cite{lindemann2021survey, ergen2019unsupervised}, and convolutional neural networks (CNNs) \cite{thill2021temporal, he2019temporal} have all demonstrated the ability model complex dependencies across variables. A review of deep learning techniques by Li and Jung (2023) \cite{li2023deep} categorizes anomalies into specific time points, intervals, or entire series. They also highlighted the need for methods that detect anomalies and explain the underlying causes.

Explainability is a critical component of anomaly detection. Understanding the drivers of anomalies is often as important as their detection. Attention mechanisms focus on the most relevant features in sequential data \cite{wang2022variational, kotipalli2024role}, providing interpretable insights into anomaly drivers through visualizations such as heatmaps. This visualization enables our understanding of the model's process. Post-hoc explainability methods, such as SHAP values \cite{lundberg2020local} and LIME \cite{garreau2020explaining}, can be integrated into anomaly detection pipelines to rank feature importance, particularly useful in high-dimensional data. Trifunov et al. (2021) \cite{trifunov2021anomaly} implemented a counterfactual reasoning approach leveraging Maximally Divergent Interval (MDI) \cite{barz2018detecting}, this method simulates "what-if" scenarios that were used for attributing anomalies to specific variables.

Multivariate climate datasets, characterized by strong temporal dependencies, require sophisticated methods to isolate the contribution of individual variables. Integrating deep learning models with domain knowledge enhances the interpretability and accuracy of anomaly attribution in climate systems \cite{jiang2024interpretable, yang2024interpretable}. Building on these advances, our research extends the application of anomaly detection and attribution to the Greenland Ice Sheet by combining counterfactual reasoning method with a Cluster-LSTM-VAE framework \cite{ale2024harnessing}. This approach identifies anomalies and attributes them to specific climate features. It provides actionable insights into surface melt dynamics at the atmosphere interface and validates the method against established techniques such as SHAP values and Random Forest-based feature importance.

\section{Methods}
\subsection{Problem Definition}
A multivariate time series (MTS) consists of multiple univariate time series \( n_i \) at the same time steps, each representing distinct climate variables. Formally, it is represented as \(x = \{X_1, X_2, X_3, \dots, X_n\}\) where \( X_t = [X_{t,1}, X_{t,2}, X_{t,3}, \dots, X_{t,n}] \) and \( t \) denotes the timestamp of the current observation. Anomaly detection aims to identify deviations from expected patterns at specific timestamps \( t_i \).

This research focuses on calculating feature attribution for anomalies detected in the MTS. Specifically, we evaluate the ability of ERA5 snowmelt and GEMB melt models to capture the dynamics of anomalous melt events. Feature attribution provides insight into the relative significance of climate variables, enabling a deeper understanding of the processes driving these anomalies.

\subsection{Cluster-LSTM-Variational AutoEncoder (CLV) Framework} \label{clv}
The Cluster-LSTM-Variational AutoEncoder (CLV) framework, introduced in \cite{ale2024harnessing}, detects anomalies in multivariate time series by combining feature clustering with an encoder-decoder architecture. We describe it here for completion to facilitate the understanding of our contributions in the instance-level attribution and comparison of the climate models. This section provides an overview of the framework, emphasizing its anomaly detection and attribution capabilities.

\noindent\textbf{Correlation-Based Clustering:}
To capture relationships among features, the input data are used to compute a correlation matrix \(R\)
\[
R = 
\begin{bmatrix} 
1 & r_{12} & r_{13} & \dots & r_{1n} \\
r_{21} & 1 & r_{23} & \dots & r_{2n} \\
r_{31} & r_{32} & 1 & \dots & r_{3n} \\
\vdots & \vdots & \vdots & \ddots & \vdots \\
r_{n1} & r_{n2} & r_{n3} & \dots & 1 \\    
\end{bmatrix} \]
where \( r_{nm} = r_{mn} \) is the Pearson correlation coefficient between variables \( X_n \) \& \( X_m \), $r_{nm} = \frac{\sum(X_n - \Bar{X}_n)(X_m - \Bar{X}_m)}{\sqrt{\sum(X_n - \Bar{X}_n)^2 \sum(X_m - \Bar{X}_m)^2}}$

We use the correlation score as a metric for feature clustering via \(k-\)means such that each variable $X_n$ is assigned to a cluster. The optimal number of clusters is determined using the Elbow method and evaluated using the Silhouette coefficient score. These clusters form the input to the CLV framework, allowing the model to encode both intra-cluster relationships and temporal dependencies.

\noindent\textbf{Encoding-Decoding for Anomaly Detection:}
The CLV framework uses an encoder-decoder architecture to learn parallel representations of each cluster and reconstruct the input data. For each time step, the encoded cluster representations are concatenated and passed to the decoder for reconstruction. Anomalies are identified by computing the reconstruction error on the test data \( X_{\text{test}} \), which contains abnormal samples. The reconstruction loss is defined as:
 
\begin{equation}
\footnotesize
\text{loss}_X = - KL(q_\phi(z|x^i) \parallel p_\theta(z)) + \mathbb{E}{q_\phi(z|x^i)}[\log p_\theta(x^i|z)]
\label{Lvae}
\end{equation}

where \( q_\phi(z|x^i) \) and \( p_\theta(z) \) represent the approximate posterior and prior distributions, respectively, and \( x^i \) denotes the input data at timestamp \( t_i \).
The loss function consists of two components: The first term, Kullback-Leibler (KL) Divergence, is a regularizer that measures the divergence between the encoder’s distribution and the prior. The second term is the Reconstruction Likelihood, which is the expected negative log-likelihood of the data point, measuring how well the input \(x^i\) is reconstructed.

The anomaly scores $X_{A_{\text{scores}}}$ are derived by combining the KL divergence and the reconstruction likelihood (Equation \ref{Lvae}). We determine anomalies due to the data being hard to reconstruct or the distribution of the latent representation of the data being far from the prior, which gives a more holistic view of the data fitting the model. A higher reconstruction loss indicates that the data point is more likely to be anomalous.

\noindent\textbf{Dynamic Thresholding:}
To adapt to localized characteristics of the data, we define a sliding window approach for temporal segmentation of the anomaly scores. The window size is determined by the attributes of the data, such as periodicity, trends, seasonality, and spatial coverage. The initial anomaly threshold is calculated for the first segment using the Peaks-Over-Threshold (POT) method, categorizing instances as normal or anomalous. The sliding window then shifts by half its span, and the threshold is recalculated for the updated segment. This dynamic thresholding approach ensures robustness across varying temporal patterns \cite{ale2024harnessing}.

By leveraging correlation-based clustering and temporal encoding, the CLV framework effectively captures both local and temporal dependencies in multivariate time series, providing a robust basis for anomaly detection and subsequent feature attribution.

\subsection{Feature Attribution} \label{feature-attri}
To determine the drivers of anomalous melt events, we build upon the CLV framework and implement a counterfactual explanation method. One way to empirically determine the cause of an event is to alter one of its potential causes at a time and observe the magnitude of change in the event. In the context of climate data, direct interventions on climate features to observe their impact on outcomes are impractical. Instead, we assess their significance by measuring their influence on the model's reconstruction loss when selectively modified.

In this method, each feature is iteratively replaced with its yearly median value to ensure perturbations remain within the realistic bounds of the data distribution. The instance-level importance of a feature is quantified as the change in reconstruction loss between the original (factual) and modified (counterfactual) instances. This change reflects the feature's contribution to deviations captured by the anomaly score.

Using this attribution method, we compare the contributions of features to anomalies detected in the ERA5 snowmelt and GEMB melt models. This comparison reveals the extent to which each model captures the dynamics of snowmelt processes within the Greenland Ice Sheet, providing insights into the individual roles of climate features in driving anomalies.

The attribution method is detailed in Algorithm \ref{alg:feature-attribution}. We represent the multivariate time series (MTS) data as a matrix $D$ of dimensions $m \times n$, where $m$ is the number of time instances and $n$ is the number of features. To calculate the anomaly score for each feature $f_i$, we construct a modified dataset $\Bar{D_i}$ by replacing the $i$-th column of $D$ with the annual median value of that column.

\[
\Bar{D_i} = 
\begin{cases} 
D & \text{if column is unaltered;} \\
\text{annual median of } D_i & \text{if column } i \text{ is replaced.}
\end{cases}
\]

Using the modified dataset $\Bar{D_i}$, we compute the anomaly scores using the CLV framework described in Section \ref{clv}.

We compare the anomaly scores across all modified features to identify the most influential feature $f_i$ for a given time instance $t$. Let $\hat{D}$ represent the matrix of anomaly scores derived from $\Bar{D}$. For each time instance $t$, we compute the set of features $F_t$, with anomaly scores exceeding the baseline anomaly score $X_{A_{\text{scores}}, t}$, as follows:
 
\begin{equation}
F_t = \{ f_i \in \Bar{D_i} \ | \ \text{anomaly score of } f_i > X_{A_{\text{scores}}, t} \}
\end{equation}

From this set $F_t$, we identify the most influential feature at time $t$ as $f_{\text{max}, t}$:

\begin{equation}
    f_{\text{max}, t} = \arg\max(F_t)
\end{equation}

The final output is a vector of length $m$, where each element corresponds to $f_{\text{max}, t}$, the feature with the greatest influence on the anomaly score at each time instance. This provides a clear and interpretable ranking of feature importance, enhancing our understanding of the dynamics driving anomalous melt events in climate models.
 
\begin{algorithm}[H]
\caption{Feature Attribution for Anomalous Melt Detection}
\label{alg:feature-attribution}
\textbf{Input:} MTS data matrix $D \in \mathbb{R}^{m \times n}$, trained \texttt{CLV\_Model}, baseline $X_{A_{\text{scores}}} \in \mathbb{R}^m$. \\
\textbf{Output:} \hspace{4.5em}$F_{\text{max}} \in \mathbb{R}^m$ for $f_{\text{max}, t}$ at each time instance $t$.\\

\begin{algorithmic}[1]
\State Initialize $F_{\text{max}} =$ [ ] 
\For{$i = 1$ to $n$} \Comment{Iterate over all features}
    \State Create modified dataset $\Bar{D}_i$:
    \State \hspace{1.2em} Replace the $i$-th column of $D$ with its annual median value.
    \State Compute anomaly scores $X_{A_{\Bar{i}}}$ for $\Bar{D}_i$ using \texttt{CLV\_Model}.
    \State Compute $\Delta X_A = X_{A_{\Bar{i}}} - X_{A_{\text{scores}}}$.
    \State Store $\Delta X_A$ for feature $f_i$.
\EndFor
\For{$t = 1$ to $m$} \Comment{Iterate over all time instances}
    \State Identify feature set $F_t = \{ f_i \ | \ \Delta X_A[f_i][t] > X_{A_{\text{scores}}, t} \}$.
    \State Find $f_{\text{max}, t} = \arg\max(\Delta X_A[f_i][t] \ \forall \ f_i \in F_t)$.
    \State Append $f_{\text{max}, t}$ to $F_{\text{max}}$.
\EndFor
\State \Return $F_{\text{max}}$
\end{algorithmic}
\end{algorithm}

\noindent\textbf{Algorithm Complexity:} In the worst-case scenario, the space and time complexity of our attribution method is linear with respect to the number of features \(n\) and time instances \(m\). This ensures scalability for moderate feature sizes but may become computationally intensive for very high-dimensional data.

\section{Experiment}
\subsection{Datasets}
This section outlines the datasets used in our experiment.
We assessed the attribution method using the WADI \cite{ahmed2017wadi} and SMD \cite{su2019robust} datasets, both of which are well-established benchmark datasets for anomaly detection. They are labeled, making them particularly suitable for evaluating and validating attribution method.
Table \ref{features} provides the key characteristics of the climate datasets used in our study, which are further described below:

\noindent\textbf{ERA5 Product \cite{hersbach2020era5}:} ERA5 is a comprehensive global climate reanalysis dataset, covering the period from January 1940 to the present. It includes a wide array of atmospheric, oceanic, and land variables, such as temperature, precipitation, and snowmelt.

\noindent\textbf{GEMB Product \cite{schlegel_2024_10806250}:} The Glacier Energy and Mass Balance (GEMB) model simulates the surface-atmospheric energy and mass exchange for ice sheets and glaciers, including the firn state. It provides data at a 5-day temporal resolution, excluding leap days, and covers the period from 1951 to 2022. Our feature of interest is the melt model.

\noindent\textbf{Derived Features:} Additional features were derived to capture relevant climate dynamics. The total wind component (\(tw10\)) was calculated as: $tw10 = \sqrt{u10^2 + v10^2}$
where \(u10\) and \(v10\) represent the 10-meter wind component. 

Similarly, the solar downward radiation absorbed by the snow (\(ssrdas\)) was computed as: $ssrdas = ssrd \times (1 - asn)$ where \(ssrd\) is the surface solar radiation downward, and \(asn\) is the snow albedo.

\noindent\textbf{MEaSUREs Greenland Surface Melt:} This data is part of the NASA Making Earth System Data Records for Use in Research Environments (MEaSUREs) program. It is an observational data of daily binary record of snowmelt on the Greenland Ice Sheet with a spatial resolution $\text{25km} \times \text{25km}$ and temporal coverage from 1979 to 2012.

\noindent\textbf{Region of Interest:} The study focuses on southwest Greenland, a region that has experienced intense surface melting in recent decades. The region spans longitude from $-52^\circ$ to $-44^\circ$ and latitude from $60.10^\circ$ to $66.85^\circ$, encompassing 924 spatial grids. For this analysis, we filtered the products to include only the months of May, June, July, August, and September for each year from 1951 to 2020, as these months typically exhibit significant melt activity.

\begin{table}[!t]
\centering
\caption{Features used in the study}
\begin{adjustbox}{max width=1\columnwidth}
\begin{tabular}{llll}
\hline
Dataset                   & Features       & Description                     \\\hline
 \multirow{10}{*}{ERA5}     & $t2m$ & 2m temperature                         \\
                            & $msl$ & Mean sea level pressure                   \\
                            & $u10$ & 10m u-component of wind                   \\
                            & $v10$ & 10m v-component of wind                   \\ 
                            & $ssrd$ & Surface solar radiation downwards      \\
                            & $strd$ & Surface thermal radiation downwards      \\ 
                            & $tcc$ & Total cloud cover                         \\ 
                            & $tp$ & Total precipitation                       \\ 
                            & $smlt$ & Snowmelt                                 \\
                            & $asn$ & Snow albedo                               \\
    GEMB                    & $gemb$ & Melt                                     \\
 \multirow{2}{*}{Derived}   & $tw10$ & Total component of wind                    \\
                            & $ssrdas$ & Solar downward radiation absorbed by the snow \\\hline 
\end{tabular}
\label{features}
\end{adjustbox}
\end{table}

\subsection{Evaluation in Benchmark Datasets}
\noindent {\bf CLV Benchmark Evaluation:} The performance of the CLV framework utilized in this research has been evaluated on benchmark datasets for anomaly detection, as detailed in \cite{ale2024harnessing}. 

\noindent {\bf Attribution Evaluation:}
To evaluate our attribution method on benchmark datasets, we utilized the WADI \cite{ahmed2017wadi} test data and the category-one (machine-1) test data from the Server Machine Dataset (SMD) \cite{su2019robust}.Using the CLV framework \cite{ale2024harnessing}, anomalies were first detected within this test data. Subsequently, our attribution method was applied to determine and rank the contribution of individual features to the detected anomalies. We identified the top 20 and top 10 features for WADI and SMD, respectively. 
These top-ranked features were then used to train a predictive binary classification model. The datasets were split into training and testing sets, and the model was trained using weighted binary cross-entropy loss. The model parameters are summarized in Table \ref{tab:model_parameters}.

\begin{table}[!t]
\centering
\caption{Model Parameters for Evaluating Feature Attribution on Benchmark Dataset}
\begin{adjustbox}{max width=1\columnwidth}
\begin{tabular}{ll}
\hline
{Parameters}         & {Values}                    \\ \hline
LSTM Layers                 & 64, 32, 10                         \\ 
Dense Layer                 & 1                                  \\ 
Dropout Layer               & 0.2                                \\ 
Activation Function         & ReLU, Sigmoid                      \\ 
Optimizer                   & Adam                               \\ 
Learning Rate               & 0.001                              \\ 
Loss Function               & Weighted Binary CrossEntropy       \\ 
Epoch                       & 50                                 \\ 
Patience                    & 10                                 \\ 
Batch Size                  & 64                                 \\ 
Window Size                 & 7                                  \\ \hline
\end{tabular}
\label{tab:model_parameters}
\end{adjustbox}
\end{table}

Our evaluation involved training the model using the top-20 features for the WADI dataset and the top-10 features for the SMD dataset, as attributed by our attribution method (discussed in Section \ref{feature-attri}). To assess the effectiveness of our unsupervised attribution method, we compared its performance against feature attributions derived from established supervised methods, including XGBoost, Random Forest, and SHAP values, using the same number of top features. Performance metrics such as precision, recall, F1 score, and PR-AUC were used for evaluation.

The results, presented in Table \ref{tab:attribution_comparison}, indicate that our unsupervised attribution method performs comparably to, and in some cases exceeds, the performance of well-established supervised methods. This demonstrates the robustness of our approach in identifying the most critical features for model training and its potential applicability in attributing climate features contributing to extreme events.

\begin{table*}[!ht]
\centering
\caption{Performance comparison of attribution methods on SMD: Top-10 and WADI: Top-20 Features.}
\label{tab:attribution_comparison}
\begin{adjustbox}{max width=1\textwidth}
\begin{tabular}{lccccc|ccccc}
\hline
{Attribution Method} & \multicolumn{5}{c|}{SMD (LSTM): Top-10} & \multicolumn{5}{c}{WADI (LSTM): Top-20} \\ 
                            & Precision & Recall & F1 Score & PR AUC & ROC - AUC & Precision & Recall & F1 Score & PR AUC & ROC - AUC \\ \hline
Ours                        & 0.9673    & \textbf{0.9660}  & \textbf{0.9667}   &0.9871 & 0.9909    & 0.9897    & 0.848  & 0.9134   & \textbf{0.9974} & \textbf{0.9599}    \\
XGB                         & 0.9544    & 0.9389 & 0.9466   & 0.9869 & 0.9930     & 0.9965    & \textbf{0.8617} & \textbf{0.9242}   & 0.9971 & 0.9548    \\
SHAP                        & \textbf{0.9679}    & 0.962  & 0.9648   & \textbf{0.9888} & 0.9927    & 0.9948    & 0.8491 & 0.9162   & 0.9969 & 0.9521    \\
RF                          & 0.9431    & 0.9457 & 0.9444   & 0.9863 & \textbf{0.9939}    & \textbf{0.998}     & 0.8134 & 0.8963   & 0.9968 & 0.952     \\ \hline
\end{tabular}%
\end{adjustbox} 
\end{table*}

\subsection{Evaluation in Climate Model Dataset}
\subsubsection{Data Preprocessing}
To begin, the GEMB melt model output was smoothed to a 0.25-degree (31 km) resolution to match the spatial resolution of the ERA5 dataset. Next, leap days were removed from the ERA5 dataset, and the time series was aggregated from daily to 5-day mean intervals to align with the temporal resolution of the GEMB melt output. 
Once both datasets were scaled to the same spatial and temporal resolution, they were merged, and a z-score normalization was applied to ensure consistency across all features. Outliers within each feature were identified using the interquartile range (IQR) method and replaced with the yearly mean of inlier values, thereby reducing irregularities and creating a smoother dataset for the reconstruction models, which perform optimally when trained on normal data. The processed data was then reshaped into a three-dimensional format of $M \times T \times F$, where $M$ represents the number of samples, $T$ the number of timesteps, and $F$ the number of features. To capture temporal dependencies, a stateful rolling window of 14 timesteps was applied to the reshaped data.


\subsubsection{Anomaly Detection Results} 
\begin{figure}[!ht]
    \centering 
    \includegraphics[width=0.85\columnwidth]{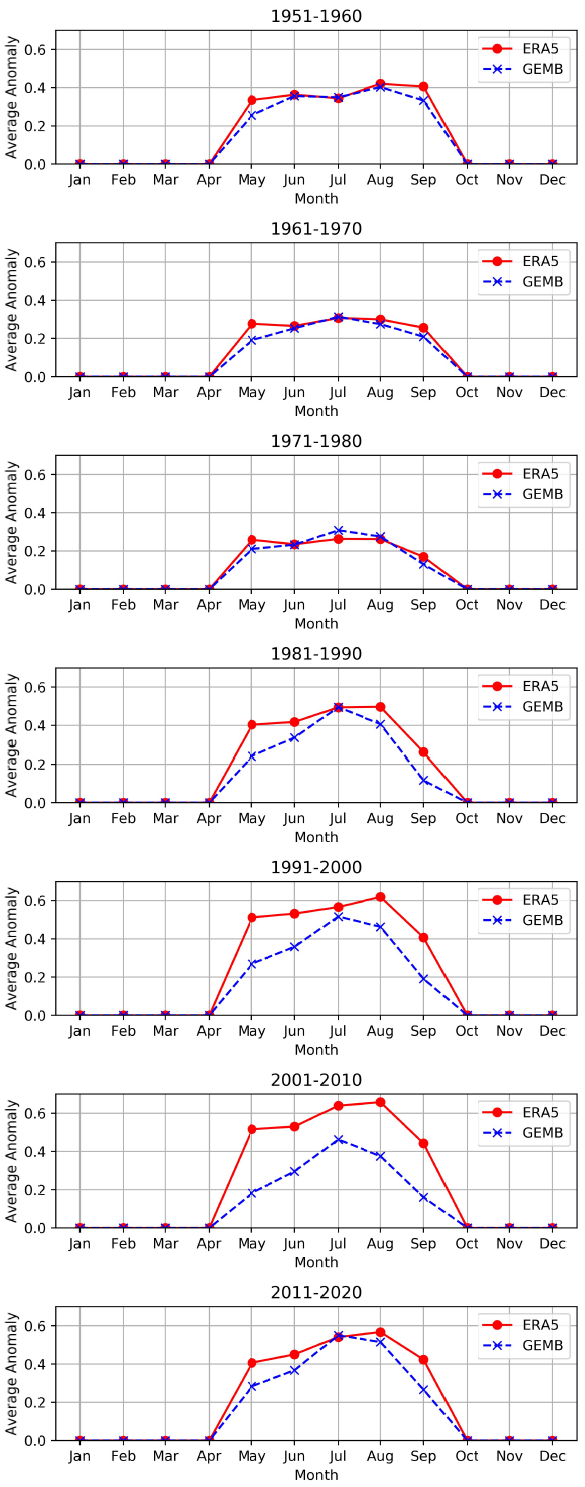}
    \caption{Decadal Analysis of the Average Monthly Anomaly Count per Grid for ERA5 and GEMB } 
    \label{monthly_agg}
\end{figure}

We conducted two sets of anomaly detection experiments: one excluding the GEMB melt model and another excluding the ERA5 snowmelt model. A decadal analysis of the monthly anomaly counts presented in Figure \ref{monthly_agg} reveals similar patterns in anomaly counts. However, the analysis using the GEMB melt model identified fewer anomalies compared to the ERA5 snowmelt model. The difference in the count of anomalies between the models could be due to their properties. The simplicity of the ERA5 snowmelt model could lead to higher detection of extreme events, potentially reflecting sensitivities in its modeling approach. Furthermore, an intensification of anomaly counts was observed in both models over time. 

To quantify the differences in anomaly counts, we performed a T-statistic test to evaluate the statistical significance of the difference in anomaly count between the number of anomalies detected with the ERA5 snowmelt model compared to the GEMB melt model, from 1981-1990 to 2011-2020. The results, presented in Table \ref{tab:t_test_results}, indicate statistically significant differences in anomaly counts for the decades 1991-2000 and 2001-2010, with p-values below the 0.05 threshold. Positive T-stat scores suggest that the mean anomaly count in ERA5 is higher than that in GEMB, reaffirming that the ERA5 snowmelt model detected more extreme events during these periods.

In addition, we analyzed the decadal increase in anomaly counts, as shown in Figure \ref{monthly_agg}. To assess the statistical significance of this trend, we applied T-statistic tests to compare anomaly counts between the first three decades (1951-1980) and the last three decades (1991-2020). The results, summarized in Table \ref{tab:era5_gemb}, indicate statistically significant increases for both models. Negative T-stat scores reveal that the mean anomaly counts in the earlier decades were lower than those in the later decades, further highlighting the intensification of anomalous events over time.

Overall, these findings highlight a clear intensification of anomalies in the Greenland Ice Sheet over time,

\begin{table}[!t]
\centering
\caption{Statistical Significance Test of Observed Differences in Identified Anomalies in ERA5 compared to GEMB } 
\begin{tabular}{lcc}
\hline
{Period} & {T-Stat} & {P-value} \\ \hline
1981-1990           & 1.3839         & 0.1677          \\ 
1991-2000           & 2.0595         & 0.0405           \\ 
2001-2010           & 3.3539         & 0.0009           \\ 
2011-2020           & 1.0743         & 0.2837           \\ \hline
\end{tabular}
\label{tab:t_test_results} 
\end{table}

\begin{table}[!t]
\centering
\caption{Statistical Significance Test of Observed Differences in Identified Anomalies Between The First and Last Three Decades in  ERA5 and GEMB }
\begin{adjustbox}{max width=1\columnwidth}
\begin{tabular}{l|cc|cc}
\hline
{Period} & \multicolumn{2}{c|}{{ERA5}} & \multicolumn{2}{c}{{GEMB}} \\ 
 & {T-Stat} & {P-value} & {T-Stat} & {P-value} \\ \hline
1951-1980 compared to 1991-2020 & -5.2913 & 1.62E-07 & -2.331 & 0.02 \\ \hline
\end{tabular}
\label{tab:era5_gemb}
\end{adjustbox}
\end{table}

\begin{table*}[!t]
\centering
\caption{Top-5 Attributed Features and Their Anomaly Attribution Frequency Across Grids (Freq = Total Count of Attributed Anomalies / The Number of Grids) for Each Decade in GEMB and ERA5 Analysis}
\label{tab:feature_counts}

\begin{adjustbox}{max width=0.95\textwidth}
\begin{tabular}{llrlrlrlrlrlrlr}
\hline
\multirow{2}{*}{Dataset} & \multicolumn{2}{c}{1951-1960} & \multicolumn{2}{c}{1961-1970} & \multicolumn{2}{c}{1971-1980} & \multicolumn{2}{c}{1981-1990} & \multicolumn{2}{c}{1991-2000} & \multicolumn{2}{c}{2001-2010} & \multicolumn{2}{c}{2011-2020} \\
 & {Feature} & {Freq} & {Feature} & {Freq} & {Feature} & {Freq} & {Feature} & {Freq} & {Feature} & {Freq} & {Feature} & {Freq} & {Feature} & {Freq} \\ \hline
{GEMB} & ssrd & 5.8247 & ssrd & 3.7035 & ssrd & 3.8041 & ssrd & 5.0628 & ssrd & 6.1396 & ssrd & 5.6255 & ssrd & 7.2251 \\ 
 & tcc & 0.9740 & tcc & 0.5465 & tcc & 0.3690 & tcc & 0.6883 & tcc & 0.9600 & tcc & 1.0574 & tcc & 1.0260 \\ 
 & ssrdas & 0.3247 & ssrdas & 0.2305 & ssrdas & 0.2587 & ssrdas & 0.2879 & ssrdas & 0.4210 & ssrdas & 0.2392 & ssrdas & 0.3712 \\ 
 & lcc & 0.3139 & lcc & 0.1742 & lcc & 0.1180 & lcc & 0.1883 & lcc & 0.2792 & lcc & 0.2597 & lcc & 0.3431 \\ 
 & tp & 0.2067 & tp & 0.1342 & tp & 0.0887 & tp & 0.1905 & tp & 0.2089 & tp & 0.2208 & tp & 0.3571 \\ \hline
{ERA5} & t2m & 5.6515 & t2m & 3.6829 & t2m & 2.7154 & t2m & 5.6959 & t2m & 7.5660 & t2m & 7.0541 & t2m & 6.4838 \\ 
 & strd & 3.6277 & strd & 2.5942 & strd & 2.0487 & strd & 4.6429 & strd & 6.1569 & strd & 6.4957 & strd & 5.3853 \\ 
 & ssrd & 3.0606 & ssrd & 2.7013 & ssrd & 2.8939 & ssrd & 4.5022 & ssrd & 5.8355 & ssrd & 6.1959 & ssrd & 4.9535 \\ 
 & tw10 & 1.0747 & tw10 & 0.7359 & tw10 & 0.6136 & tw10 & 0.6753 & tw10 & 1.2197 & tw10 & 1.5942 & tw10 & 1.5703 \\ 
 & ssrdas & 0.5119 & v10 & 0.4762 & v10 & 0.4026 & v10 & 0.6937 & v10 & 0.9123 & v10 & 1.2229 & v10 & 0.7760 \\ \hline
\end{tabular}%
\end{adjustbox}
\end{table*}

\subsubsection{Anomaly Attribution} 
Using the method described in Section \ref{feature-attri}, we analyzed feature attributions for anomalies detected by the GEMB and ERA5 models across 924 spatial grids for each decade. The results, summarized in Table \ref{tab:feature_counts}, present the average number of extreme events attributed to each climate feature. This average is calculated as the total count of attributed anomalies divided by the number of grids, providing insight into the frequency with which specific features contribute to extreme event detection.

For ERA5, the most frequently attributed features were $t2m$, $strd$, $ssrd$, $tw10$, and $v10$. 
These results align with the physical processes underlying the ERA5 snowmelt model's surface energy balance framework, where temperature ($t2m$) and radiative processes ($strd$ and $ssrd$) are central to snowmelt dynamics. Notably, an increase in the frequency of these features over the last three decades reflects the documented intensification of temperature across the Greenland Ice Sheet. This trend underscores the critical role of temperature changes in driving anomalies in the region.

In contrast, GEMB's feature attribution reveals a different set of significant contributors to the detected anomalies. $ssrd$ emerges as a prominent feature with a markedly higher anomaly count in the last three decades compared to the first three. Other significant features include $tcc$, $ssrdas$, $lcc$, and $tp$. This broader and more diverse attribution reflects the complex surface snow dynamics captured by the GEMB melt model, which extends beyond surface energy balance to incorporate processes such as snow accumulation, firn densification, ice melt, and meltwater percolation to depth. The observed increase in anomaly counts for $ssrd$ and $tcc$ highlights the strong influence of solar radiation and cloud cover on the cryosphere's energy balance. The physical processes of both models captured by our analysis validate the attribution method. Note that the very high attribution of $ssrd$ could be due to the ERA5 snow albedo used.

The decade of 2011–2020 witnessed two of the most extreme melt events recorded in the Greenland Ice Sheet. During this period, GEMB attributed higher anomaly counts to $tp$, $lcc$, and $ssrd$, highlighting the role of precipitation and radiative processes in enhancing melt dynamics. This indicates GEMB's ability to capture interactions between precipitation, cloud cover, and solar radiation, which are critical to understanding the drivers of extreme melt. In contrast, ERA5 showed a slight decrease in anomaly counts, possibly reflecting its reliance on a simplified energy imbalance framework that may not fully account for the complex interplay of precipitation and cloud-related processes with surface snow dynamics on ice sheets.

The consistent increase in anomaly counts across features and decades supports the conclusion that dynamic surface melt in southwest Greenland is intensifying. This analysis reveals distinct differences in feature attribution between the GEMB and ERA5 models. These differences highlight the inherent variations in how the models interpret and represent the melt dynamics on the ice sheet surface. GEMB's attribution results suggest a focus on complex interactions involving features such as $ssrd$, $lcc$, and $tp$, which are directly tied to melt dynamic processes beyond the surface energy balance. In contrast, ERA5 emphasizes features such as $t2m$, $strd$, and $ssrd$, which are closely related to its energy imbalance-driven snowmelt model. These contrasting attributions underscore the complementary strengths of the two products in capturing the drivers of extreme events.

\subsubsection{Ground Truth Validation}
\begin{figure*}[!ht]
    \centering  
    \includegraphics[width=1\linewidth]{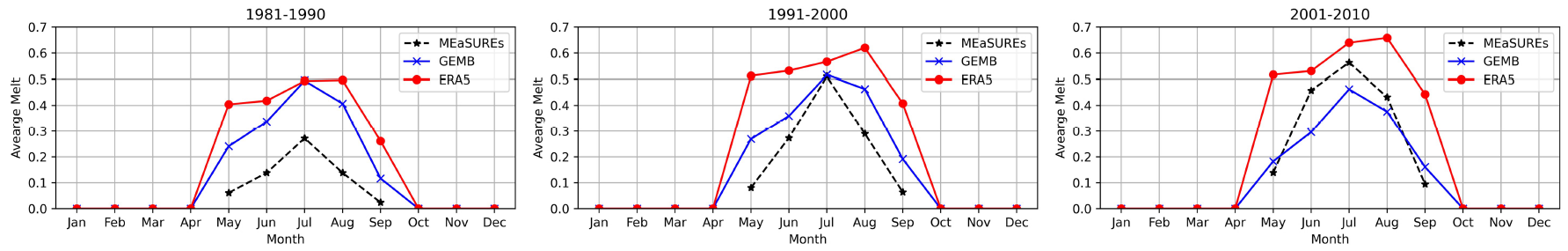} 
    \caption{Observed monthly average melt per grid for MEaSUREs data and the monthly average anomaly per grid for ERA5 and GEMB for the period 1981-1990 to 2001-2010}  
    \label{fig:measures-label}  
\end{figure*}

To evaluate the alignment between modeled anomalies and observed melt patterns, we compared the monthly average occurrence of snowmelt from the MEaSUREs dataset per grid (Figure \ref{fig:measures-label}) with the monthly average identified anomalies per grid derived from ERA5 and GEMB (Figure \ref{monthly_agg}). The results reveal a closer similarity between the GEMB model and the MEaSUREs ground truth data.
In ERA5, we observed an increase in anomalies detected during the transitional months of May and September. The increase may indicate a heightened sensitivity of ERA5 to increasing temperatures during these months. However, this trend is less pronounced in GEMB, which may indicate that while temperatures are rising, they have not yet consistently translated into significant melt events during these months. The differences in these models compared to MEaSUREs ground truth data might be an overestimation of melt anomalies or other feature anomalies during the transitional months.


\begin{table*}[!t]
    \centering
    \caption{Comparison of Our Findings with Previous Literature}
    \begin{adjustbox}{max width=1\textwidth}
    \begin{tabular}{p{2.5cm}p{7cm}p{7cm}}
    \hline
    {Findings} & {Our Findings} & {Previous Literature}\\ \hline
    
    
    {Decadal Increase in Anomalies} 
    & A statistically significant increase in the identified anomalies was observed when comparing the first three decades to the last three decades. We have a t-test score of -5.2913, p-value of 1.62E-07 for ERA5 product and -2.331 t-test score, p-value of 0.02 for GEMB product.
    & Multiple studies have identified an increase in extreme events recorded in the Greenland Ice Sheet. For instance, Bonsoms et al. \cite{bonsoms2024rising} stated that ice and snow melting events have become about twice as frequent during recent summer decades compared to the period 1950-1990. Other relevant works include \cite{slater2021increased, beckmann2023effects}. \\ \hline
    
    {Attribution Reflects Model Physics} 
    & Our attribution method successfully identified features that reflect the distinct physical properties of ERA5 snowmelt and GEMB melt model.  
    & Previous studies have noted that ERA5 snowmelt calculations are fundamentally based on the energy balance between incoming (solar radiation) and outgoing energy at the snow surface, with a positive imbalance leading to melting \cite{krebs2023brief}. In contrast, the GEMB model incorporates additional processes, including shortwave subsurface penetration, meltwater retention, percolation, refreeze, snow grain size, dendricity, sphericity, and compaction \cite{gardner2023glacier}. \\ \hline
    
    {Key Features of Attribution} 
    & ERA5 attribution highlights an intensification in temperature-related features, with $t2m$, $strd$, and $ssrd$ being the most attributed, aligning with the energy balance mechanisms of ERA5 snowmelt. Conversely, GEMB attribution emphasizes the cryospheric dynamics of the region, identifying $ssrd$, $lcc$, and $tp$ as the most significant features. Reduced cloud cover increases solar radiation reaching the ice sheet, enhancing melt. 
    & Hofer et al. \cite{hofer2017decreasing} attributed Greenland’s declining surface mass balance not to direct responses from temperature increases or longwave radiation anomalies but to reduced cloud cover and increased shortwave radiation anomalies. Similar conclusions were drawn by \cite{blau2024extreme, ryan2024contribution, wang2021greenland}. \\ \hline
    
    \end{tabular}
    \label{tab:comparison_findings}
    \end{adjustbox}  
\end{table*}

Other findings are categorized into three distinct subject areas, summarized, and validated against previous ground-truth studies. Table \ref{tab:comparison_findings} compare our findings with the existing literature, highlighting the alignment and insights gained. For instance, 
the increase in decadal anomalies corroborates broader trends reported in studies such as Bonsoms et al. \cite{bonsoms2024rising}, highlighting a doubling of extreme melt events in recent decades.

Our attribution analysis also demonstrates the ability of our method to reflect the physical characteristics of the respective models (ERA5 and GEMB). ERA5 primarily attributes anomalies to temperature-related features, consistent with its energy imbalance framework, while GEMB attribution incorporates more complex cryospheric dynamics. These observations are supported by the work by Gardner et al. \cite{gardner2023glacier}, which describes the advanced processes modeled in GEMB. In general, this validation underscores the significance of our approach in accurately identifying features and trends within climate data. It reinforces the applicability of the proposed methods in understanding surface snow and ice dynamics on ice sheets.
 
\section{Significance}
This research underscores the necessity of incorporating complex physical models beyond surface energy balance approaches in modeling melt processes. Doing so provides a more accurate attribution of future anomalies and a deeper understanding of the underlying drivers of surface melt. The analysis demonstrates that our feature attribution method captures the physical processes represented by climate models, offering insight into what these models are simulating. This capability bridges the gap between model predictions and physical interpretation, enhancing the interpretability of climate models. Scientists could analyze and predict extreme melt events by identifying how individual variables contribute to melt dynamics. Furthermore, these insights will support decision-makers in managing and mitigating climate-related impacts in the Arctic while contributing to global climate adaptation and mitigation strategies.

\section{Conclusion}
This research presents a comprehensive analysis of the ERA5 snow-\\ melt and GEMB melt models, evaluating their capabilities to capture surface melt dynamics within southwest Greenland. Using an unsupervised anomaly detection framework (CLV) enhanced with feature attribution techniques, we identified key climate features driving melt anomalies and assessed the distinct strengths and limitations of each model. The results highlight that the ERA5 snowmelt model, based primarily on energy imbalance calculations, detects a higher frequency of extreme melt events, potentially due to its sensitivity to temperature-driven processes. In contrast, the GEMB melt model, which incorporates complex cryospheric processes such as snow accumulation, firn densification, and meltwater refreezing, provides a broader representation of the physical dynamics underlying melt. Both models revealed a statistically significant increase in anomaly frequency over the last three decades, indicating an intensification of extreme events within the Greenland Ice Sheet.

The analysis highlights a notable increase in anomaly frequency during the shoulder months of May and September for ERA5 compared to GEMB. This underscores the differences in the models' sensitivity and representation of early—and late-season dynamics. ERA5 anomalies were mainly attributed to temperature-driven features such as $t2m$, $strd$, and $ssrd$, consistent with its energy-balance approach. GEMB, on the other hand, emphasized features like cloud cover ($lcc$) and solar radiation absorbed by the snow ($ssrdas$), reflecting its capacity to model complex interactions influencing melt processes. These differences underscore the importance of selecting a model based on the study's specific objectives.

Our proposed unsupervised attribution method effectively identified critical drivers of anomalies, aligning with the physical processes underpinning each model. This alignment validates the robustness of the method for anomaly detection and feature attribution in multivariate climate data. While the results align with the expected physics of the two models, they emphasize the utility of our approach for evaluating and interpreting climate models in complex environments. Future work will extend this analysis to the entire Greenland Ice Sheet and incorporate additional data sets to further validate the findings. Expanding the study will provide a broader perspective on melt dynamics and offer deeper insights into the evolving impacts of climate change on polar regions.

In conclusion, this research underscores the value of advanced machine learning techniques in understanding and monitoring ice sheet surface melt dynamics. By providing actionable insights into the drivers of extreme events, this study contributes to polar science and climate change discussions, informing strategies to monitor and mitigate the impacts of a warming climate.

\section*{Acknowledgment}
This work is funded by the National Science Foundation (NSF) Award \#2118285.
The WADI dataset was provided by iTrust, Center for Research in Cyber Security, Singapore University of Technology and Design.

\bibliographystyle{ACM-Reference-Format}
\bibliography{reference}

\end{document}